\documentclass[12pt]{elsarticle}

\usepackage{amssymb, amsmath}
\usepackage{multirow}
\usepackage{rotating}
\pdfoutput=1

\journal{Signal Processing}

\begin{document}

\begin{frontmatter}



\title{Full-reference image quality assessment by combining global and local distortion measures}


\author{Ashirbani Saha$^\dag$, Q. M. Jonathan Wu}
\address{Department of Electrical and Computer Engineering,\\
University of Windsor,\\
Canada N9B 3P4}
\address{email:\{sahai,jwu\}@uwindsor.ca}
\address{$^\dag$Tel: 1-519-253-3000, Fax: 1-519-971-3695}

\begin{abstract}
Full-reference image quality assessment (FR-IQA) techniques compare a reference and a distorted/test image and predict the perceptual quality of the test image in terms of a scalar value representing an objective score. The evaluation of FR-IQA techniques is carried out by comparing the objective scores from the techniques with the subjective scores (obtained from human observers) provided in the image databases used for the IQA. Hence, we reasonably assume that the goal of a human observer is to rate the distortion present in the test image. The goal oriented tasks are processed by the human visual system (HVS) through top-down processing which actively searches for local distortions driven by the goal. Therefore local distortion measures in an image are important for the top-down processing. At the same time, bottom-up processing also takes place signifying spontaneous visual functions in the HVS. To account for this, global perceptual features can be used. Therefore, we hypothesize that the resulting objective score for an image can be derived from the combination of local and global distortion measures calculated from the reference and test images. We calculate the local distortion by measuring the local correlation differences from the gradient and contrast information. For global distortion, dissimilarity of the saliency maps computed from a bottom-up model of saliency is used. The motivation behind the proposed approach has been thoroughly discussed, accompanied by an intuitive analysis. Finally, experiments are conducted in six benchmark databases suggesting the effectiveness of the proposed approach that achieves competitive performance with the state-of-the-art methods providing an improvement in the overall performance.
\end{abstract}


\end{frontmatter}


\section{Introduction}
\label{sec:Intro}
Automatic evaluation of the perceptual quality of a distorted image with respect to the original high quality image is technically called full-reference image quality assessment or FR-IQA. Though methods for evaluating subjective image quality by directly employing human observers are available, they are time consuming and less economic compared to an automatic evaluation using an FR-IQA technique. Given the applicability of FR-IQA in image acquisition, watermarking, restoration, fusion etc., it has gained immense popularity and attention since the last decade. There are FR-IQA techniques that employ the properties of Human Visual System (HVS) directly, and there are others, which have hypotheses supporting some properties of HVS. Some methods of FR-IQA are combined approaches involving both of the aforesaid techniques~\cite{WangIQABook}. Though peak-signal-to-noise ratio (PSNR) follows none of the approaches, it is one of the best perceptual evaluators for mainly content-independent noise~\cite{WangIWSSIM11, Zhang11}. Still, the general applicability of PSNR is very low when considered with other types of distortions or across several types of distortions. This is where the development and enhancement of FR-IQA techniques became necessary~\cite{WangIQABook}. Therefore, from the last decade, a number of different techniques has been developed for FR-IQA. We have the popular SSIM~\cite{WangSSIM04}, which is based on the property of HVS to extract structural similarity. Several SSIM based or inspired approaches and analysis~\cite{Silvestre-BlanesSP11} have been developed since then. MSSSIM~\cite{WangMSSIM04} incorporates some flexibility to SSIM by embedding multiscale information in it. Also, information content weighted SSIM, called IWSSIM~\cite{WangIWSSIM11} uses pooling based on information content calculated using mutual information from three pairs of images selected from reference, distorted, perceived reference and perceived distorted images. However, the first methods to use mutual information in FR-IQA are Information Fidelity Criterion (IFC)~\cite{SheikhIFC05} and Visual Information Fidelity (VIF)~\cite{SheikVIF2006}. Hence, they are called information theoretic approaches.

Among other approaches are phase congruency based technique feature similarity (FSIM/FSIM$_c$ for color images)~\cite{Zhang11} and its modifications to form phase deviation sensitive energy features based similarity technique (PDSESIM) for FR-IQA~\cite{SahaPDSESIM13} to improve the general applicability. In a similar framework, a faster technique called spectral residual based similarity (SR-SIM)~\cite{ZhangSRSIM12} has been developed and shown to perform well in three databases. SR-SIM uses saliency maps derived from spectral residual technique to find perceptual similarity between pixels as well as to pool the quality map obtained from the reference and distorted images. The work presented in~\cite{SahaSaliencyIQA13} points out the parameter dependence of the said framework by experimental analysis. A number of techniques show interesting steps towards separating distortions in an image according to some pre-defined criteria. Most Apparent Distortion (MAD)~\cite{Larson10} segregates the distortions present in an image as near threshold or supra-threshold and uses dual strategies for evaluating those distortions. In the same context, another approach is based on defining the distortions as additive or detail loss based and combining their measurements (ADM)~\cite{LIAIMDLM11}. A recent approach~\cite{JwuIGM2013} uses the principal of internal generative mechanism (IGM) of human brain and dissolves the given images into predicted and disorderly parts. Then the distortions on these two parts are separately evaluated by different techniques and combined to form the objective score. MAD, ADM and Visual-Signal-to-Noise ratio (VSNR)~\cite{ ChandlerVSNR07} employ the direct applications of the HVS properties like contrast sensitivity and visual masking. Most of the methods mentioned in the previous discussions form the state-of-the-art techniques of FR-IQA.
The existence of so many techniques validates the general observation that each of them has some shortcomings for certain distortions and databases on which they are evaluated. An FR-IQA technique that performs well for all degradations (provided in the benchmark databases) does not exist. Thus, development of an FR-IQA technique that performs well with all possible types of distortions is still an open problem.

As already discussed, MAD, ADM and IGM  apply different strategies to separately evaluate different aspects of distortion. Inspired by all of these, the aspect we focus on is the automatic evaluation of image quality based on global and local perceptual visual cues and their combination. As explained in~\cite{CogPsychoBook}, visual processing is a simultaneous combination of bottom-up and top-down processes. Bottom-up processing focuses on highlighting the relatively important regions in an image~\cite{KochSaliency1998}. On the other hand, top-down processing caters to goals and it searches actively for local features~\cite{OlivaTopDownSal} based on contextual information. In HVS, these processes occur simultaneously and quickly. Whenever, a human subject is given the task of evaluating perceptual quality, he has the final goal to rate the distortions present. Based on this, we hypothesize that three main events taking place during this evaluation: 1. Initially, the global image content determines the varying attention (regionwise) of the evaluator; 2. existing regional distortions are used to assign local quality evaluations; 3. finally, based on the attention, a refinement on the previous evaluation is carried out and final rating is provided. To serve the first event, we resort to the global perceptual features of the image. For the second step, local attributes between the reference and test/distorted images are compared. Finally, third step involves a pooling strategy taking place based on the global perceptual features. In the present work, the perceptual quality of an image is expressed in terms of image saliency maps (as global perceptual features) obtained from bottom-up saliency models, gradient information and local standard deviation (local features) also termed as RMS (root-mean-squared) contrast. The saliency map of an image represents global information about how often a particular pixel is gazed at. On the other hand, local standard deviation and gradient information signify the local and contextual information of any pixel. The local correlation between the global information obtained from the reference and distorted images is computed. The local correlation between the gradient information obtained from the reference and distorted images are also calculated. These local correlations are combined with the local RMS contrast between the images. From the experimental results, we find that the integration of simple visual details of global perceptual difference information and local information may result in an effective FR-IQA technique. It differs from its predecessors in terms of treatment of local and global features by using the regional correlation. It has been shown in~\cite{LiuIQA2012}, gradient is structure-variant as well as contrast-variant. Thus similar variations in gradient magnitude values and standard deviation are expected for a pixel. However, change in standard deviation may not be caused by change in gradient magnitude only. Gradient orientation is also affected by presence of distortion. Thus, the proposed approach applies all of these visual details to arrive at the quality score. The performance analysis of the technique in six benchmark databases shows the promise of the proposed method as a competitive technique in FR-IQA.
Also, we carry out analysis on distortion dependent performance of the FR-IQA techniques using color based representation. This representation of the results clearly shows that with certain distortions, the FR-IQA techniques fail to perform well. The representation also depicts the competitive performance of the proposed method as also analyzed in Section~\ref{ssec:ExptAnDistwise}.

The remaining parts of the paper are arranged in the following manner. The motivating factors for the proposed approach are presented in Section~\ref{sec:MotivatingFactors}. The details of the proposed method are presented in Section~\ref{sec:ProposedMethod}. The performance of the proposed method is analyzed in Section~\ref{sec:ExperimentsAnalysis}. The concluding remarks are presented in Section~\ref{sec:Conclusion}.

\section{Motivating Factors}
\label{sec:MotivatingFactors}
The main motivation behind the proposed approach is to successfully combine simple visual cues, representing the global and the local information present in an image, to formulate a competitive FR-IQA technique. We have selected saliency maps as global perceptual features for an image. For local features, block based gradients and standard deviations are used. The proposed technique obtains global distortion information by comparing global features obtained from the reference and distorted images using local correlation. The local distortion information is obtained by comparing local features using local correlation and local difference. In the following sections, we describe the importance of each of the global features and the roles they are expected to play as parts of an FR-IQA technique. Based on their properties, we hypothesize that the effective combination of these simple features can result in a competitive FR-IQA technique.
\subsection{Global Features}
\label{ssec:globalfeatures}
The saliency map of an image is chosen as a representation of the global perceptual features of an image. Since saliency maps have perceptual information contained in them, they have been used as features and for pooling purposes~\cite{ZhangSRSIM12, MaSalIQAICIC08}. Given saliency maps, it becomes easier to point out the pixels which are perceptually more important than the others as they are more significantly gazed at. Many saliency methods rely on bottom-up (task-independent and data driven) processing only. Some techniques also apply top-down (task-driven, prior knowledge) approaches to enhance saliency~\cite{Fang2012}. A comparison of several types of spectral saliency based FR-IQA is presented in~\cite{SahaSaliencyIQA13}. We chose the bottom-up technique for saliency calculation called spectral residual (SR)~\cite{HouSaliency} and phase Fourier transform (PFT)~\cite{GuoPQFT08} to be used in our method owing to their simplicity, fast computation, high average performance and applicability with both grayscale and color images~\cite{SahaSaliencyIQA13}. Thus, we present two versions of the proposed method using SR and PFT separately. These versions are named as global local distortion with SR (GLD-SR) and global local distortion with PFT (GLD-PFT).

\begin{figure}
\begin{center}
\includegraphics[width = \linewidth]{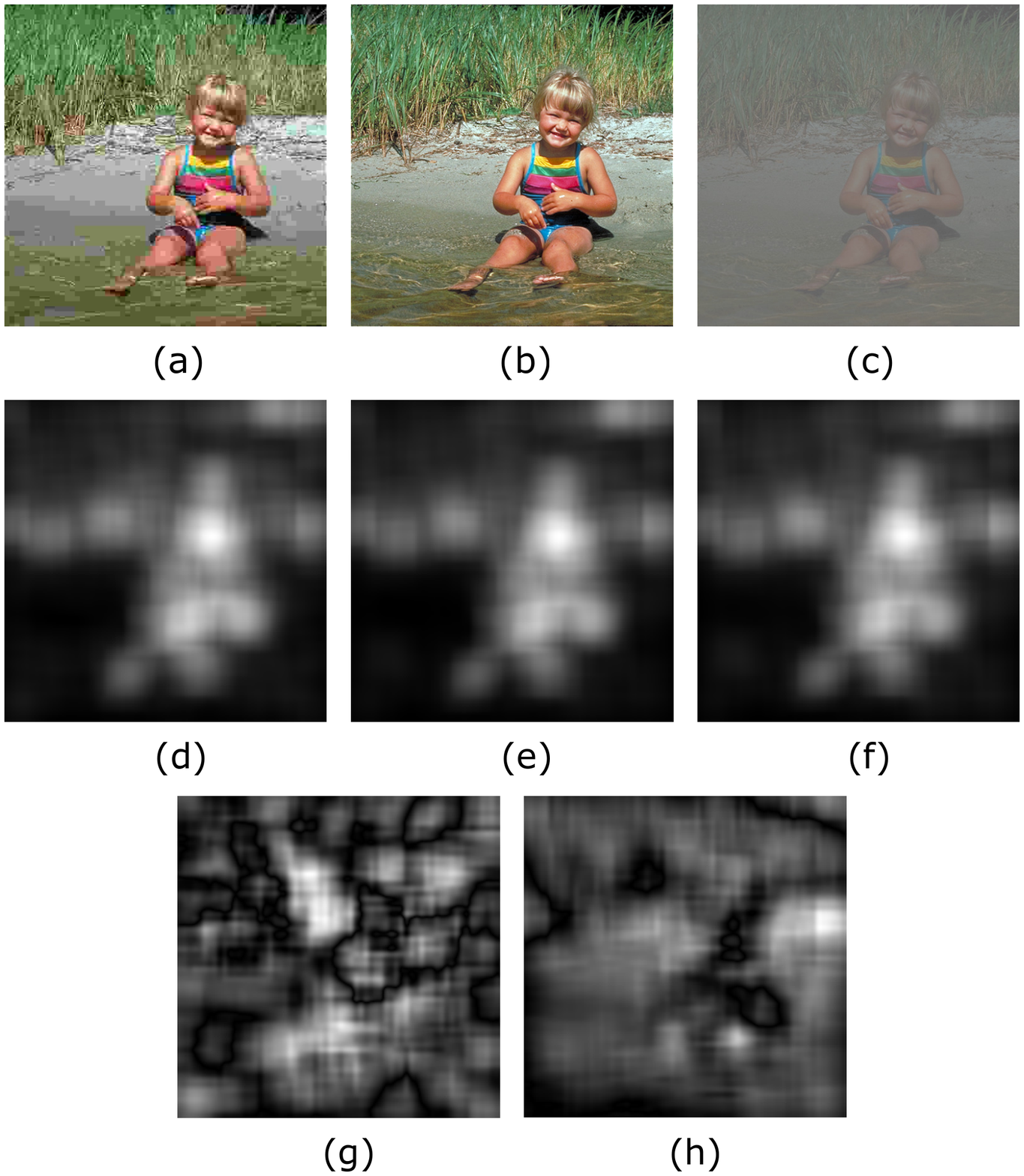}
\end{center}
\caption {\label{fig:saliencyExample}(a) and (c) are the jpeg compressed and global contrast decremented images obtained from the reference image (b). The saliency maps of (a), (b) and (c) are shown in (d), (e) and (f) respectively. The difference of the saliency maps shown in (e) and (d) is (g). The difference of the saliency maps shown in (e) and (f) is (h). }
\end{figure}

The saliency maps shown in the corresponding images as global features during our discussion of the motivation, are computed using SR only. As shown in Figs.~\ref{fig:saliencyExample}(d--f), saliency maps of different images (see Figs.~\ref{fig:saliencyExample}(a--c)) having different distortions look similar. Though the points of gazes remain similarly distributed visually,  subtle changes in values occur as shown in the Figs.~\ref{fig:saliencyExample}(g--h)). Hence, the saliency maps do not vary in similar way for different distortions. Though, the face of the girl remains among the most salient parts in all the images, the changes in saliency values at various pixels are different. Thus, global perceptual information varies for different distortions and this change is conveyed by the saliency maps.

This property of the saliency maps is used in the proposed technique. The local correlations of the saliency maps are expected to be higher if the variations of the saliency values in the block-wise neighborhood of the same pixel are similar. However, this similarity indicates similar variation in the global perceptual information for the neighborhood but same values of saliency may not be present in both of the corresponding blocks of the images. Also, local distortions may be ignored by the global representation. Hence, the use of local visual cues remains important. Local standard deviation and gradient information are used as local visual information and these are elaborated in the following section.

\subsection{Local Features}
\label{ssec:localfeatures}
\begin{figure}
\begin{center}
\includegraphics[width = \linewidth]{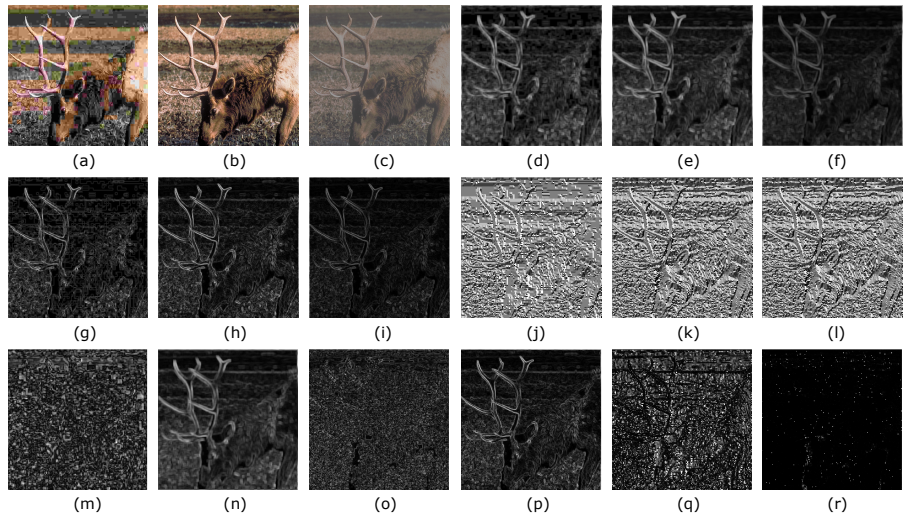}
\end{center}
\caption {\label{fig:varianceGradient}In the first row, (a) and (c) are the jpeg compressed and global contrast decremented images obtained from the reference image (b). The RMS contrast images corresponding to (a), (b), (c) are presented in (d), (e) and (f) respectively. The second row represents the corresponding gradient magnitude images (g, h, i) and the gradient orientation images (j, k, l). In the third row (m), (o), (q) represent the difference images calculated between the RMS contrast, gradient magnitude and orientation images of the reference and jpeg compressed images, respectively. (n), (p), (r) represent the difference images calculated between the RMS contrast, gradient magnitude and orientation images of the reference and global contrast decremented images, respectively.}
\end{figure}

As experimentally validated in the work of Bex and Makous~\cite{Bex02}, RMS contrast can be used to detect contrast changes. Therefore, RMS contrast has been chosen to extract local features for the proposed work. As per the definition of RMS contrast, it is similar to the standard deviation of luminance values. In Fig.~\ref{fig:varianceGradient}, we show the RMS contrast maps for the reference and distorted images. Some structure differences are present between the images in Figs.~\ref{fig:varianceGradient}(d) and (e) as revealed in their difference image Fig.~\ref{fig:varianceGradient}(m). On the other hand, the dynamic range of RMS contrast is lower in Fig.~\ref{fig:varianceGradient}(f) compared to that in Fig.~\ref{fig:varianceGradient}(e) and hence their difference shown in Fig.~\ref{fig:varianceGradient}(n) bears much similarity with Fig.~\ref{fig:varianceGradient}(e). The gradient magnitude image in  Fig.~\ref{fig:varianceGradient}(g) shows the outlines of the blocks of image shown in Fig.~\ref{fig:varianceGradient}(a). The difference of gradient magnitude of the jpeg compressed image and that of the reference image is shown in Fig.~\ref{fig:varianceGradient}(o). This difference reveals smaller structural changes compared to Fig.~\ref{fig:varianceGradient}(m). For the global contrast decremented image, the difference in gradient magnitude image Fig.~\ref{fig:varianceGradient}(p) reveals magnitude changes (lower dynamic range) but this image is much sharper than Fig.~\ref{fig:varianceGradient}(n). The gradient orientation image in Fig.~\ref{fig:varianceGradient}(j) has changed significantly from Fig.~\ref{fig:varianceGradient}(k). For jpeg compressed image, significant change in gradient orientation can be seen through the difference image shown in Fig.~\ref{fig:varianceGradient}(q). However, for contrast decremented image, the change in gradient orientation is very less as shown in Fig.~\ref{fig:varianceGradient}(r). Thus, the RMS contrast, gradient magnitude and gradient orientation maps have different visual depictions for different distortions as shown in the third row of the Fig.~\ref{fig:varianceGradient} and each of these local features has its own relevance.

\subsection{Correlation between the Feature Maps}
\label{ssec:localfeatures}
\begin{figure}
\begin{center}
\includegraphics[width = \linewidth]{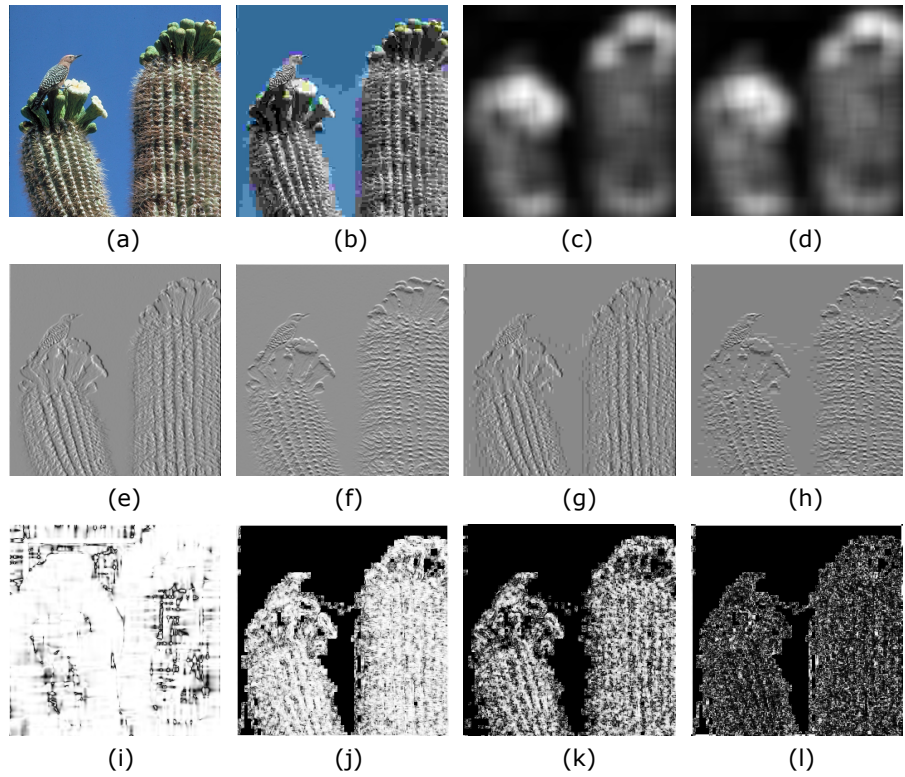}
\end{center}
\caption {\label{fig:CorrelationImportance}In the first row, (a) and (b) are the source and the jpeg compressed images; (c) and (d) are the respective saliency maps; (e) and (g) are the respective x - gradient maps while (f) and (h), the respective y-gradient maps. (i) is the block-wise correlation between (c) and (d); (j) and (k) are formed using the maximum and minimum of the block-wise correlation between (e) and (g) and that between (f) and (h). (l) represents the pixelwise differences between (j) and (k).}
\end{figure}

We have used cross-correlation between the saliency maps, x-partial derivative maps and y-partial derivative maps to find the relative local variations of the selected features. The cross-correlation conveys whether the local and global information we employ in our technique, cause uniform or varying changes throughout different neighborhoods of the pixels. A block-wise correlation between the saliency maps is sure to indicate how similarly the saliency varies in the reference and the distorted images in the same neighborhood/region. Thus, we have the idea, from global perspective, of how the saliency varies in respective local neighborhoods of a pixel. Similarly, for the x and y-partial derivatives of the local data, we get the varying trends in neighborhoods along the x and y directions. Any edge present in an image is sure to have a vertical and a horizontal component. These components are the x and y-partial derivatives. Thus, the correlation between x-partial derivatives of the reference and distorted images will depict the variation in neighborhood due to changes in edge strengths in vertical direction. Similarly, in horizontal direction, the correlation between y-partial derivatives is important. In Fig.~\ref{fig:CorrelationImportance}, the importance of correlation maps are shown for the distortion jpeg-compression. The block-wise correlation between the x-partial derivatives (Figs.~\ref{fig:CorrelationImportance}(e) and (g)) is found out for the reference and distorted images (Figs.~\ref{fig:CorrelationImportance}(a) and~\ref{fig:CorrelationImportance}(b), respectively). Similarly, the block-wise correlation between y-partial derivatives (Fig.~\ref{fig:CorrelationImportance}(f) and (h)) is calculated. Therefore, two correlation maps are obtained. The maximum and minimum of these correlation maps are used to form the images shown in Figs.~\ref{fig:CorrelationImportance}(j) and (k), respectively. The difference between these two images is shown in Fig.~\ref{fig:CorrelationImportance}(l) and it shows to what extent the local data based maximum and minimum correlation varies. This figure depicts all the edges of blocks formed in the distorted image and hence is indicative of some structural changes.

\section{Proposed Method}
\label{sec:ProposedMethod}
\begin{figure}[h!]
\begin{center}
\includegraphics[width = \linewidth]{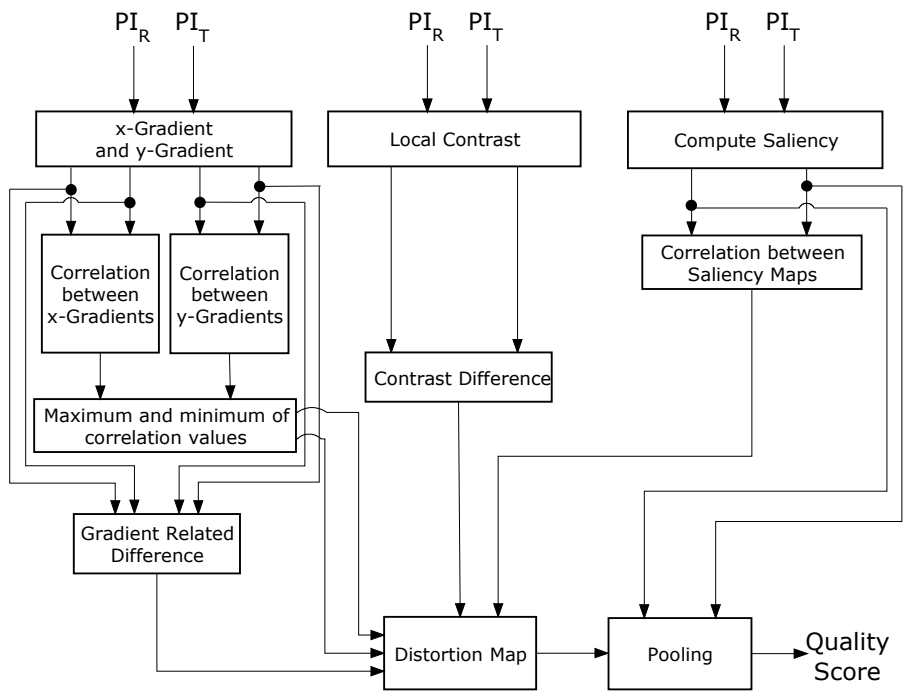}
\end{center}
\caption {\label{fig:proposedMethodBlockDiagram}A schematic diagram of the proposed method}\vspace{-.5cm}
\end{figure}

We describe the proposed method in details in this section. We sub-divide the section into three parts. The first part describes the computation of global and local feature information from the distorted images. The second part describes the use of these information to form distortion map and computation of the quality score while the third part analyzes the method using an example. The graphical depiction of the proposed method is presented in Fig.~\ref{fig:proposedMethodBlockDiagram}.

\subsection{Global and Local Feature Extraction}
\label{ssec:extractFeature}
The reference and distorted/test images $I_{R}$ and $I_{T}$ are pre-processed to obtain the images $PI_{R}$ and $PI_{T}$ respectively. The pre-processing steps are: (1) any color image is converted to grayscale; (2) the grayscale image is subjected to the process of automatic scale selection as mentioned in~\cite{WangViewpoint} and (3) the range of image intensity values is restricted between 0 and 1. In the next step, we calculate the saliency maps, gradient maps and local contrast maps. The saliency maps ($S_{R}$ and $S_{T}$) of $PI_{R}$ and $PI_{T}$ are found by spectral residual method~\cite{HouSaliency}. The local contrast is calculated as
\begin{equation}
\label{eq:RMScontrast}
V_{R}(p) = \left[\frac{1}{MN}\sum\limits_{\nu\in\omega}(PI_{R,\nu} - \mu_\omega)^{2}\right]^{(0.5)},
\end{equation}
where $\mu_\omega$ is the mean of the block $\omega$ of size $M\times N$ surrounding the pixel $p$ of image $PI_{R}$. Similarly, one can compute the local contrast image $V_{T}$ from $PI_{T}$. Then, the local contrast difference is found out as
\begin{equation}
\label{eq:RMScontrastDiff}
LC_{d}(p) = \left[\frac{(V_{R}(p) - V_{T}(p))}{2}\right]^{2}.
\end{equation}
For the local contrast and gradient calculation, we use a $3\times3$ window.
The x-gradient ($G_{R,x}$ and $G_{T,x}$) and y-gradient ($G_{R,y}$ and $G_{T,y}$) images are found by the Scharr gradient operator~\cite{ScharrOperator}. Next, we find the local correlation image $SM_{c}$ between the global saliency maps. From the gradient images, the gradient magnitudes are found out as
\begin{subequations}
\begin{align}
        G_{R,M}(p) = ({G^2_{R,x}(p)} + {G^2_{R,y}(p)})^{\frac{1}{2}},\\
        G_{T,M}(p) = ({G^2_{T,x}(p)} + {G^2_{T,y}(p)})^{\frac{1}{2}}.
\end{align}
\end{subequations}
The gradient orientations  are calculated as
\begin{subequations}
\begin{align}
        G_{R,O}(p) = \arctan({G_{R,y}(p)} / {G_{R,x}(p)}),\\
        G_{T,O}(p) = \arctan({G_{T,y}(p)} / {G_{T,x}(p)}),
\end{align}
\end{subequations}
such that the angles lie within $[-\pi,\pi]$. We calculate the gradient related difference as
\begin{equation}
G_{d}(p) = \left[\frac{\max\left({\frac{|G_{R,M}(p) - G_{T,M}(p)|}{\sqrt{2}},\frac{|G_{R,O}(p) - G_{T,O}(p)|}{2\pi}}\right)}{2}\right]^{2}
\end{equation}
which considers the maximum of the differences of gradient magnitudes and gradient orientations by using the function max(.). Since, the intensity values in the images are restricted between 0 and 1, the gradient magnitudes can have a maximum value of $\sqrt{2}$. Hence, these magnitudes are divided by $\sqrt{2}$ such that their values remain below 1. Again, the maximum difference between the orientation angles can be $2\pi$. Therefore, the orientation difference is divided by $2\pi$ such that its maximum value is restricted to 1.
\subsection{Formation of Distortion Map}
\label{ssec:distortionMap}
In order to form the distortion map first, the three local cross-correlation maps are found out between the pair of saliency maps, x-partial derivative maps and y-partial derivative maps. Thus, we have $SM_{c}$, $X_{c}$ and $Y_{c}$ denoting the local correlation maps corresponding to saliency maps, x-gradient maps and y-gradient maps respectively. Now $SM_{c}$ is the correlation between the global features, and high value of this correlation implies greater global similarity between the images. High values of correlation for the local information imply similar local variation between the images. The maximum and minimum values of correlation between the local features for any pixel $p$ are computed as $H_{c}(p) = \textrm{max}(X_{c}(p),Y_{c}(p))$ and
$L_{c}(p) = \textrm{min}(X_{c}(p),Y_{c}(p))$ respectively. Now, we calculate the primary distortion measure for each pixel $p$ as
\begin{equation}
\label{eq:MoreCorrWithGlobDataP1}
D_{p}(p) = \frac{\max(|H_{c}(p) - L_{c}(p)|,(1 - X_{c}),(1-Y_{c}),(1-SM_{c}))}{2}\times T(p),
\end{equation}
where
\begin{equation}
\label{eq:MoreCorrWithGlobDataP2}
T(p) = \left[LC_{d}(p)\times\frac{(1-SM_{c}(p))}{2}\times G_{d}(p)\right]^{\frac{1}{3}}.
\end{equation}
Therefore, $D_p$ indicates a measure of difference between two pixels considering their local neighborhood as well as global perceptual importance. The first difference term within the function max(.) in Eqn.~\ref{eq:MoreCorrWithGlobDataP1} shows the difference in correlation caused by varying changes in the neighborhoods along the horizontal and vertical directions. The
correlation differences between the local contrast and gradient difference are also used to account for the local contrast and gradient changes. The difference in local contrast and gradients are also considered to find their variation together. The correlation ranges from -1 to 1. Hence, when any correlation term is subtracted from 1, the maximum value it can take is 2. Hence, a division by 2 is carried out. Thus, the maximum deviation of any type of correlation is captured in this equation. Regarding the calculation of $T(p)$ which uses the change in local distortion obtained from the RMS contrast and gradient difference, if there exists high correlation within the saliency maps, the difference/distortion values get effectively lowered. Therefore, multiplication by $(1-SM_{c}(p))$ has been used. Hence, the product of these three terms are taken in Eqn.~\ref{eq:MoreCorrWithGlobDataP2}. The values of $T(p)$ will be larger when aforesaid differences are higher along with smaller values of $SM_{c}(p)$ (smaller values of $SM_{c}(p)$ imply significantly different local variations in global information of the reference and distorted image). $T(p)$ is multiplied by the maximum change captured through correlations calculated from global and local features. In this primary map, the treatment with all pixels remains the same. However, this map can undermine local variations in images as it relies more on the global variations.

Therefore, one more map is used based on the relative values of the correlation between the local and global data. Higher correlation in the global features compared to the local features does not indicate that the distortion is low, but local variations in luminance/RMS contrast should be regulated or considered, if needed. Therefore, the following map is formed for some selected pixels which are more correlated globally than locally. We find the set $\eta_G$ of those pixels $p$ for which the correlation between the global features is greater than that between local features, meaning $SM_{c}(p)>L_{c}(p)$. The distortion measure for them is calculated in two ways: first, the changes in RMS contrast are regulated and next, only the local differences are considered. For the first part,
\begin{equation}
\label{eq:MoreCorrWithGlobData_A}
A(p) = \begin{cases}\left(LC_{d}(p)\times\frac{(1-SM_{c}(p))}{2}\right)^{\frac{1}{2}}, \textrm{ if } p\in\eta_G\\
       0 \textrm{, otherwise}\end{cases}.
\end{equation}
This equation says that if the correlation between global features is higher than the minimum local correlation obtained from gradients, the local contrast difference has to be given varied importance as decided by $(1-SM_{c}(p))$.

On the contrary, local distortions present in some pixels in set $\eta_G$ may be significant if the local changes are high . Hence, the product of RMS contrast difference and gradient difference are considered for the second part.
\begin{equation}
\label{eq:MoreCorrWithGlobData_B}
B(p) = \begin{cases}\left(LC_{d}(p)\times G_{d}(p)\right)^{\frac{1}{2}} \textrm{ if } p\in\eta_G\\
       0 \textrm{, otherwise}  \end{cases}.
\end{equation}
The final distortion map is calculated as
\begin{equation}
\label{eq:MoreCorrWithGlobData_D_f}
D_{f}(p) = D_{p}(p) + A(p) + B(p).
\end{equation}
Finally, the quality score $Q$ is calculated as
\begin{equation}
\label{eq:MoreCorrWithGlobData_Q_s}
Q = k\frac{\sum\limits_{p=1}^{N_{I}}D_{f}(p)\times\textrm{max}(S_{R}(p), S_{T}(p))}{\sum\limits_{p=1}^{N_{I}}\textrm{max}(S_{R}(p), S_{T}(p))},
\end{equation}
where $N_{I}$ is the total number of pixels in $PI_{R}$ and saliency map values are used for the pooling the map. $k$ is a constant taken as 10000 in all our experiments. It scales the objective scores which are otherwise very low. Multiplication by this constant does not have impact on the quantitative results rather it makes the interpretation of the scores easier. In the next section, we analyze the proposed method using the $Q$ values obtained on a suite of distorted images.

\subsection{Analysis Using an Example}
\label{ssec:distortionMap}
\begin{figure}
\begin{center}
\includegraphics[width = \linewidth]{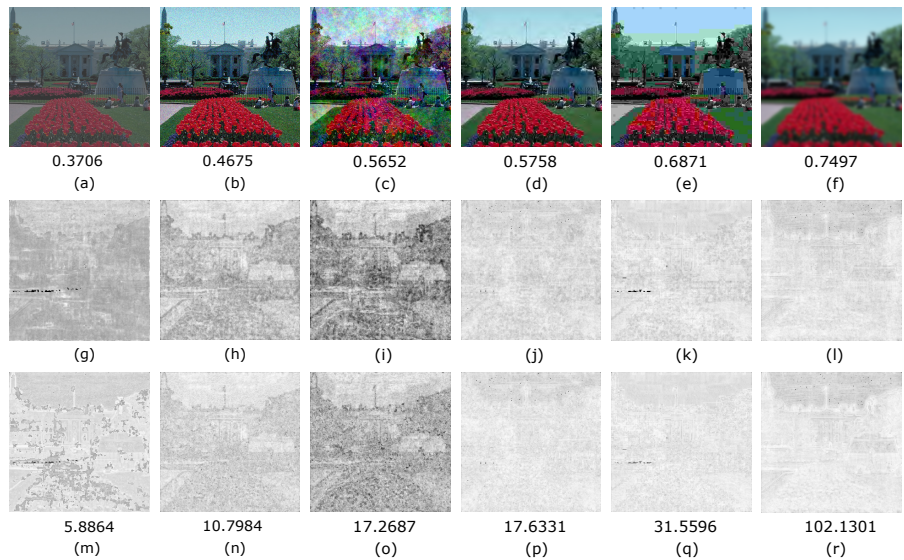}
\end{center}
\caption {\label{fig:proposedMethodAnalysis}The objective values obtained by the proposed method on six images of CSIQ database. The images (from left to right) in the first row have different types of distortions: contrast decrement, awgn, fnoise, jpeg2000 compression, jpeg compression and blur. The images are arranged in increasing order of their DMOS scores and hence perceptual quality decreases from (a) to (f). The respective DMOS values are shown below the figures in first row. The middle row shows the corresponding primary distortion maps. The final row shows the saliency weighted final distortion map which is used to calculate the objective score $Q$ using GLD-SR. The respective $Q$ values are shown below the figures in the last row.}
\end{figure}

In the diagram shown in Fig.~\ref{fig:proposedMethodAnalysis}, we find the distorted images with their corresponding primary distortion maps and saliency weighted final distortion maps computed using GLD-SR. The dynamic ranges of these maps have been adjusted to increase visual details. The brighter an area is in the distortion map, the higher is its contribution in the distortion map. The distortion map indicates the assessment of degradation of the image prior to the pooling process.  Also, this figure shows the difference between the primary and the final distortion maps. For example, the relative contrast of the values is lower in the final map than the primary map. This is as if taking `one more look' at the local distortions before evaluating the image. Hence, the final distortion map is necessary to calculate. The averages of the saliency weighted final distortion maps (the $Q$ values) in Figs.~\ref{fig:proposedMethodAnalysis}(m--r), have increased gradually demonstrating the increasing degradation of perceptual quality as we move from left to right. As the images are arranged in increasing order of their DMOS values (provided in the CSIQ database), the objective scores should be in the same order as well. Since the proposed technique is a way to quantify the perceptual degradation, values of $Q$ increase as perceptual degradation increases. The values of $Q$ are found to be increasing in the same order as the DMOS values. We find that the proposed technique is able to maintain the same order of perceptual quality as given by human observers for different types of degradations. We find that only parameter the proposed technique has, is the window size which is fixed for all our experiments. Increasing window size has no significant influence on the performance but it increases the computational time.

\section{Experiments and Performance Analysis}
\label{sec:ExperimentsAnalysis}
We are presenting the experiments and results of our proposed method in this section. Extensive experiments are carried out in six popular IQA databases: LIVE database~\cite{SheikDB}, CSIQ database~\cite{Larson10}, TID2008 database~\cite{PonomarenkoTID2008}, A57 database~\cite{ChandlerVSNR07}, MICT database~\cite{ToyamaDB} and IVC database~\cite{IVCDB}. We have used five evaluation measures to compare the FR-IQA techniques as inspired by the works of VQEG~\cite{VQEG2003}. These evaluation measures are: Spearman's Rank-Order Correlation Coefficient (SROCC), Kendall's Rank-Order Correlation Coefficient (KROCC), Pearson's Linear Correlation Coefficient (PLCC), Mean Absolute Error (MAE) and Root Mean Square Error (RMSE). A five parameter logistic function~\cite{Sheik06} is used for mapping between the subjective scores of the images given in the databases and the corresponding objective scores for the calculation of PLCC, MAE and RMSE. The closer the values of SROCC, KROCC and PLCC are to 1, the better is the FR-IQA technique. In this work, SROCC values below 0.8 have been considered weak correlation and those above 0.95 are considered strong and used for further analysis in Section~\ref{ssec:ExptAnDistwise}. On the other hand, for MAE and RMSE smaller value indicate better FR-IQA technique. In addition to these, we have carried out the F-test to analyze the statistical significance of the proposed method. The proposed methods GLD-SR and GLD-PFT have been compared with 12 other state-of-the-art FR-IQA techniques: PSNR, SSIM, MS-SSIM, VIF, VSNR, MAD, IWSSIM, FSIM$_c$, ADM, SR-SIM, IGM and PDSESIM. We have gradually moved from global representation of experiments and results to a much detailed one. In the first experiment, the average performance of the techniques across all distortions and databases are considered. In the next experiment, we look at individual performance of the FR-IQA techniques in 6 databases considering all distortions in a database together, along with a statistical analysis using F-test. Finally, we look at the distortion wise analysis of the results in three largest databases. This three-fold experimentation helps to deeply analyze the proposed method by several aspects and shows its competitiveness to the state-of-the-art techniques.

\subsection{Average Performance Comparison}
\label{ssec:ExptAverage}
The average performance of all the methods in the six databases is shown in Table~\ref{Tab:DBAvg} using SROCC, KROCC and PLCC. Average results with these three evaluation measures are considered due to the following reason. As per convention the absolute values of these measures (SROCC, KROCC, PLCC) are taken. Therefore, their values are bounded by 0 and 1. However, for MAE and RMSE, the values are affected by the range in which subjective human ratings provided in each database lie. Hence, their lower bound is 0 but upper bound is dependent on subjective score ranges. Therefore, average values of MAE and RMSE may be misleading.

The direct average for each evaluation measure shown in Table~\ref{Tab:DBAvg} is the arithmetic mean of the corresponding measure computed from six databases. The weighted average of an evaluation measure is the weighted mean of the measure values and the number of images in a database acts as its weight. The top performances for each evaluation measure in direct and weighted average are represented in bold font.We find that the proposed method GLD-SR has the average values in terms of all of the three measures for direct average. GLD-PFT has close performance with PDSESIM using SROCC and KROCC. The average PLCC score for GLD-PFT is better than all existing techniques and slightly below GLD-SR. For weighted average, GLD-PFT is the best followed by GLD-SR and PDSESIM using all of the three correlation scores. Thus, with the proposed methods GLD-SR and GLD-PFT, we have been able to achieve improved results compared to the state-of-the-art techniques across six databases considering all distortions together.

\subsection{Database wise Performance Comparison}
\label{ssec:ExptAverage}
In this experiment, the overall performance of an FR-IQA technique in each of the six databases is evaluated and the results are shown in Table~\ref{Tab:DBEvalAll}, using all five evaluation measures. The top three performances for each evaluation measure in each database are highlighted in bold. As we can find, none of the FR-IQA techniques are among the top three in all databases. VSNR, IWSSIM, ADM, SR-SIM and IGM are among the top performers in only one of the databases. MAD, PDSESIM and the proposed approaches GLD-PFT and GLD-SR are among the top performers in three to four databases; but the subset of these databases varies in each case. For the CSIQ and IVC databases, the proposed approach GLS-PFT gives best performance using all of the five evaluation measures. The performance of GLD-SR is better than the existing FR-IQA techniques in CSIQ, TID2008 and IVC databases using some of the evaluation measures. In TID2008 database, GLD-SR and GLD-PFT perform better than several state-of-the-art FR-IQA techniques using some of the evaluation measures.

\begin{figure}[h!]
\centering
\includegraphics[width = \linewidth]{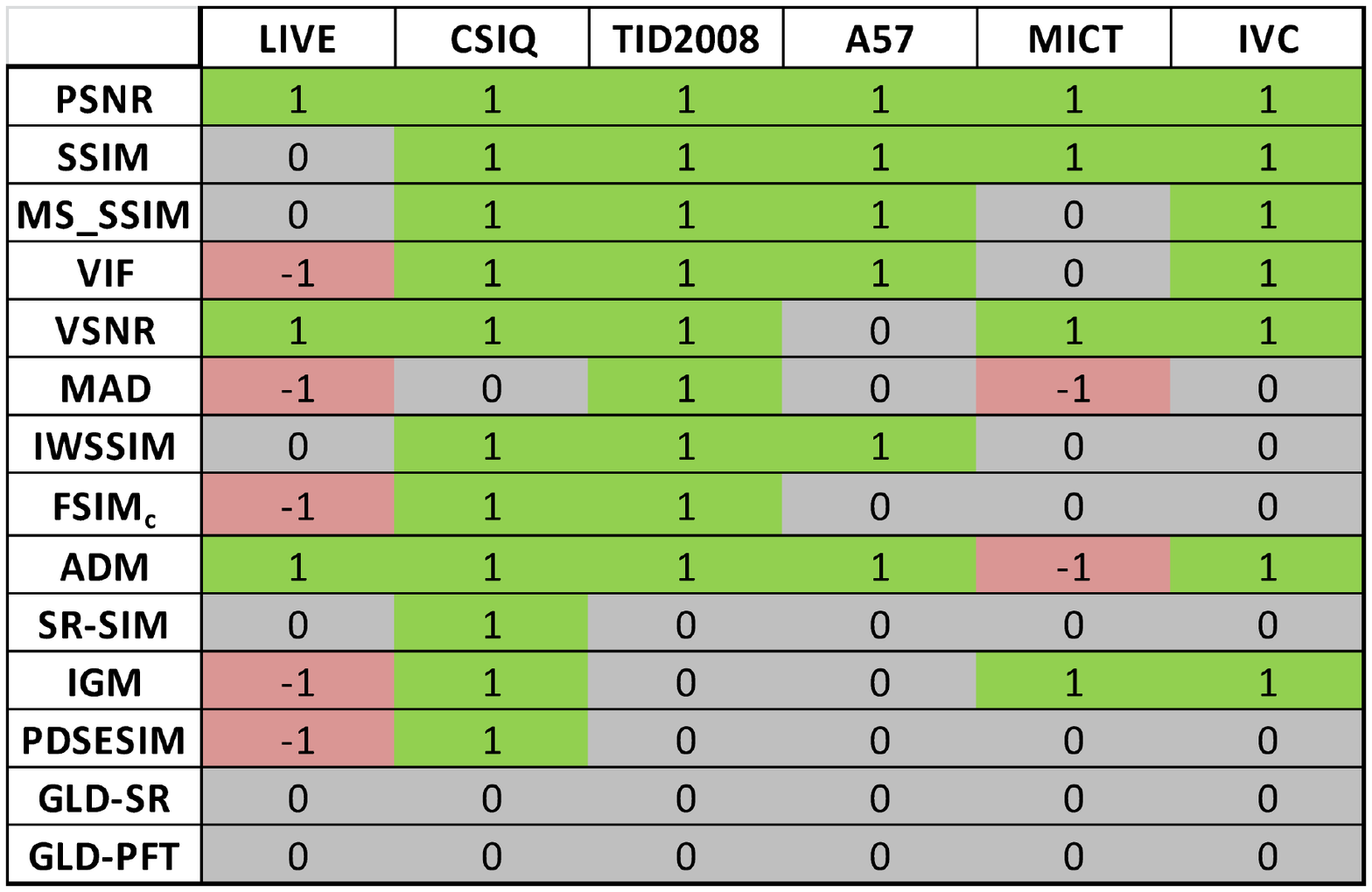}
\caption{F-ratio test for comparing GLD-SR with 13 other FR-IQA techniques in 6 databases.}
\label{Fig:FRatioResults_SR}
\end{figure}

\begin{figure}[h!]
\centering
\includegraphics[width = \linewidth]{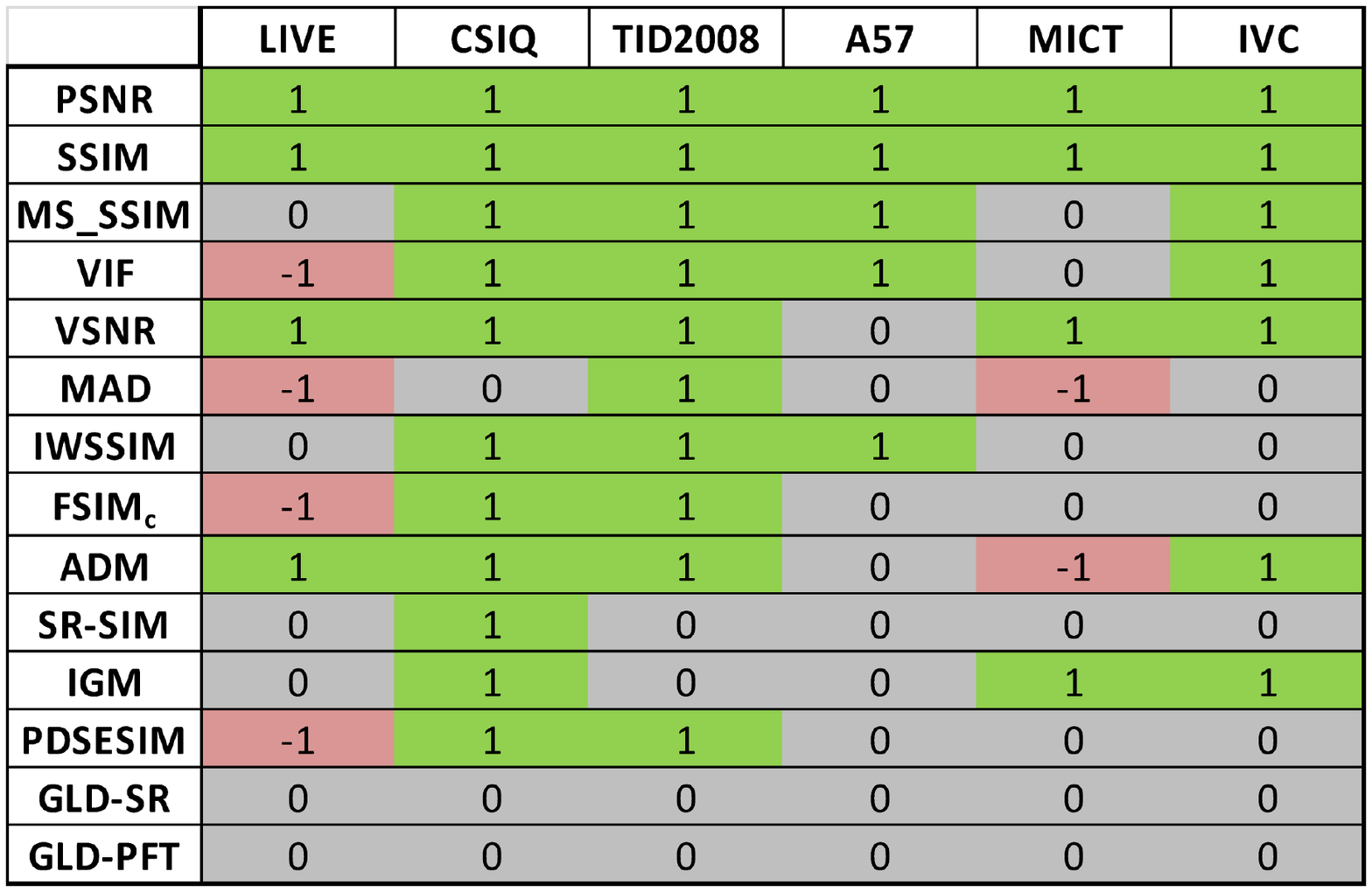}
\caption{F-ratio test for comparing GLD-PFT 13 other FR-IQA techniques in 6 databases with}
\label{Fig:FRatioResults_PFT}
\end{figure}

To analyze the statistical effectiveness of GLD-SR and GLD-PFT, F-test has been used. The F-test is carried out using the residuals of the objective scores (after mapping) and the given subjective scores. The variance of the residuals for each database is calculated for all the FR-IQA techniques. Depending on the number of images in each database, the critical value of F-distribution $F_{c}$, is determined at 95$\%$ confidence level. For a given database, if the ratio of the larger variance to the smaller one is greater than $F_{c}$, the FR-IQA method with the smaller variance is significantly better than the one with larger variance. The result of the F-test is shown in Figs.~\ref{Fig:FRatioResults_SR} and ~\ref{Fig:FRatioResults_PFT}. Fig.~\ref{Fig:FRatioResults_SR} shows the statistical comparison of GLD-SR with other 13 techniques. Each cell in the figure with a numeric value has values containing `-1',`0' or `1' only. `0' implies that GLD-SR is not statistically different from the corresponding FR-IQA technique given in the row heading for the database provided in the column heading. `1' implies that GLD-SR is better than the FR-IQA technique in the given databases indicated by the row and column headings respectively. Finally, `-1' implies significantly worse performance.  We have also colored the cells depending on the its values (-1 : pink, 0 : grey, 1 : green) to make the comparison vivid. From the F-test results shown in Fig.~\ref{Fig:FRatioResults_SR}, we understand that the more number of green cells signifies better performance of GLD-SR. Since, comparison of an FR-IQA technique with itself will return a `0', the row corresponding to GLD-SR is totally grey. Hence, the maximum number of green cells expected is 13*6 = 78. The number green cells obtained is 40. Hence, in 51.28\% of the cases, GLD-SR has shown improvement. Similarly, GLD-PFT has improved in 52.56\% of the cases as shown in Fig~\ref{Fig:FRatioResults_PFT}. While some performance decline is exhibited in the LIVE and TOY databases, in majority of the databases (CSIQ, TID2008, A57 and IVC) performance has improved.

\subsection{Distortion wise Performance Comparison}
\label{ssec:ExptAnDistwise}
\begin{sidewaysfigure}
\vspace{-5cm}
\centering
\includegraphics[width = \linewidth]{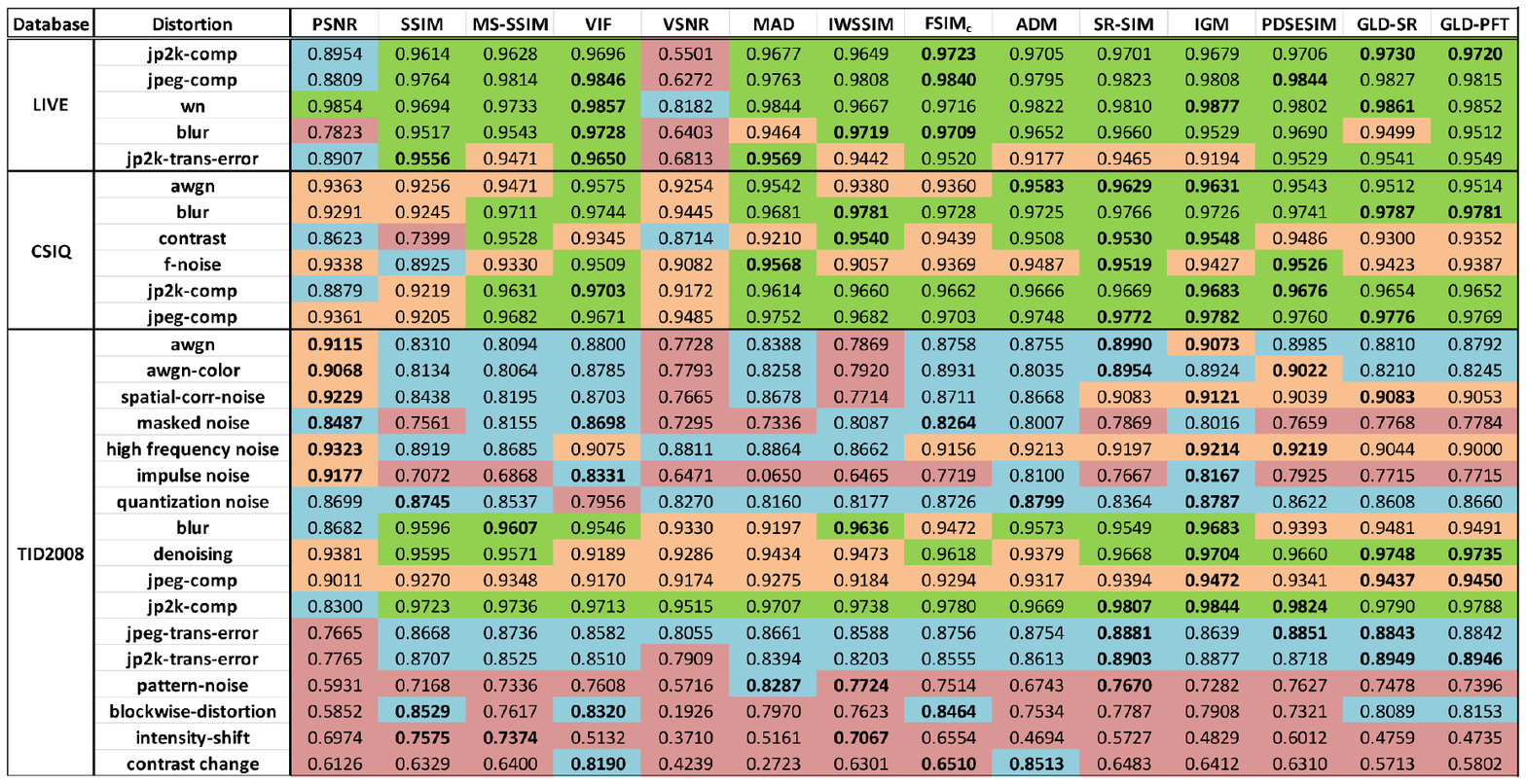}
\vspace{-7cm}
\caption {Distortion wise SROCC scores in LIVE, CSIQ and TID2008 databases. }
\label{Fig:DBDetails}
\end{sidewaysfigure}

In our last experiment, three databases (LIVE, CSIQ, TID2008) with the highest number of distorted images are chosen. In each of these databases, more than 4 distortions exist. Using SROCC, the performance of GLD-SR and GLD-PFT in comparison with several existing FR-IQA techniques for different distortions is demonstrated in Fig.~\ref{Fig:DBDetails}. The top three performers for each distortion are highlighted in bold. An interesting point to be noted here: 13 methods (out of 14) have been among the top assessment methods for at least one distortion. Thus, this figure gives an insight about the suitable methods, if the distortion is fixed. VIF and GLD-SR are among the top methods for 9 distortions whereas, SR-SIM and IGM are among top methods for 10 and 14 distortions respectively. For several types of noises in the TID2008 database, PSNR performs better than the other techniques.

Apart from this, each cell with a numeric value is highlighted with a specific color depending on which group the value falls within. To elaborate, the background color in each cell depends on the value contained in the cell. As per the convention, absolute values of SROCC are shown and divided in four ranges. Hence, the range of absolute SROCC values [0,1] is divided in four non-overlapping groups: group 1 [0 - 0.7999] is represented by pink, group 2 : [0.8 - 0.8999] is presented by blue, group 3 [0.9 - 0.9499] is presented by orange and group 4 [0.95 - 1] is presented by green. The grouping is made non-uniform so as to segregate the strong performances in each distortion from the moderate and weak performances.
First, we take a look at the results of TID2008 database. For the distortions like pattern noise, block-wise distortion, intensity shift and contrast change, the SROCC values for all of the techniques are very low. On the other hand, most of the techniques have higher SROCC scores for distortions like blur, denoising, jp2k-comp and jpeg-comp. PSNR performs much better than most of the techniques in awgn, awgn-color, spatial-corr-noise, masked noise, high frequency noise, impulse noise and quantization noise. For jpeg-trans-error and jp2k-trans-error we have mixed results from the techniques but a strong performing techniques is still missing. On the other hand, in CSIQ database, some techniques have lower SROCC values for contrast but other distortions are well dealt by most of them. Only SR-SIM is found to have strong performance in all distortions in CSIQ database. Most techniques have high SROCC scores for all distortions in LIVE database except for VSNR and PSNR. Also, SSIM, MS-SSIM, VIF, FSIM$_{c}$, PDSESIM and GLD-PFT have top performances for all distortions in LIVE database as indicated by the green color. Most techniques perform very well in LIVE database. Except for f-noise in the CSIQ database, all distortions present in the TID2008 database are present in LIVE and CSIQ databases and most techniques have high SROCC values for the common distortions present in LIVE, CSIQ and TID2008 databases. However, the percentage of green cells shows a huge decline when LIVE and CSIQ databases are compared with the TID2008 database.

From the above discussion, one can conclude that there is no assurance of better overall performance in a database, if the distortion wise performance is better. This is because, relative assessment of different distortions will be required together for overall performance. If the objective scores obtained from different distortions have different ranges, the overall performance in the database is likely to be poor. As we have seen, MAD has the best overall performance in LIVE database. Though MAD has strong performance in four out of five distortions in LIVE database while several others techniques have five strong performances, it still performs best when all of the distortions are considered together. A closer look at the results show that the proposed techniques follow a similar trend. GLD-SR and GLD-PFT have strong performance in CSIQ database for four out of six distortions. However, their overall performance is better than SR-SIM (6 strong performances), IGM (5 strong performances) as well as MAD, VIF, IWSSIM and ADM in CSIQ database. Hence, both overall and distortion wise performance analysis are necessary to assess an FR-IQA technique.

\subsection{Discussion on the parameters}
\label{ssec:distortionParameters}
A look into the details of the methods reveals of the adjustable parameters present in the models. For PSNR, the definition is conventional and no adjustable parameters are involved. For SSIM, window size is one of the parameters and based on it, the structure, contrast and luminance similarity maps are generated. This way of generating similarity maps is followed in FSIM$_{c}$ and SR-SIM as well. It is shown in~\cite{SahaPDSESIM13} that this framework introduces parameters based on the number of similarity maps calculated and hence, increases the effective number of adjustable parameters for the method. MS-SSIM has parameters similar to SSIM and the number of scales used also affects. However, number of scales and scale wise weights are fixed for MS-SSIM. For VIF, different types of modeling techniques are employed and in course of that, window size, number of wavelet scales used and noise variance of modeling neural distortions are introduced as the adjustable parameters. MAD uses log-Gabor decompositions in which the scales and orientations need to be pre-defined while VSNR has the number of scales as the only adjustable parameter. ADM uses wavelet decomposition to separate the image into two parts using 4 scales to compute the necessary parts. IGM uses auto-regression based models to generate the orderly and disorderly portions. The window size is a parameter as well as another parameter is introduced by the constant required in the luminance comparison maps. The saliency based method SR-SIM has window size and constants required by the similarity maps as the adjustable parameters. Similarly, for PDSESIM, tunable parameters are introduced by its similarity calculation framework.

Based on the discussion presented, we find that apart from PSNR, the parameter dependency is evident among the existing FR-IQA techniques. The proposed method is dependent on the saliency map but it does not compute the similarity map unlike SR-SIM and PDSESIM. Hence, it has no tunable parameters apart from the window size. A window size is required to calculate the local correlation between the maps and it is same as the window size used to calculate the local standard deviation and gradient information.
\section{Conclusion}
\label{sec:Conclusion}
In this paper, we have tried to approach the problem of FR-IQA by hypothesizing about the simultaneous top-down and bottom-up processing that take place when a human observer rates the subjective quality of an image. Global perceptual information of the image stands for the bottom-up processing. On the other hand, local visual cues as gradient information and local standard deviation are used to imitate the top-down processing which is affected by the goal of the observer to rate the image quality. The regional correlation of the global information between the reference and test images and the same from the local information are used to compute the global and local distortion values. These values are combined to arrive at the final degradation score. Experiments carried out in six databases validate the promise of the proposed approach. The promising performance of the proposed method can be attributed to combination of local and global distortion measure which is a collection of simple visual cues proven to be effective from the existing IQA research. At the same time, this method also follows the drawback of the existing FR-IQA techniques to evaluate the certain types of distortions with lesser accuracy. Thus, the future work will be aimed at the distortion independent performance of the FR-IQA techniques. Also, in the proposed approach, the luminance component has only been used to formulate the technique. In future, we would study the effect of the chrominance related information on the quality score.
\section*{Acknowledgment}
This work is supported in part by the Canada Research Chair Program and the Natural Sciences and Engineering Research
Council of Canada.

\bibliographystyle{elsarticle-num}
\bibliography{refsSept2014}

%


\setlength{\tabcolsep}{1.2pt}
\renewcommand{\arraystretch}{0.7}
\begin{sidewaystable}\footnotesize
\begin{center}
\caption{\label{Tab:DBAvg}Average Performance in 6 databases}
\begin{tabular} {cccccccccccccccc}
\hline
\hline
\bf{Method}& \bf{Measure}&\bf{PSNR}&\bf{SSIM}&\bf{MS-SSIM}&\bf{VIF}&\bf{VSNR}&\bf{MAD}&\bf{IWSSIM}&\bf{FSIM$_c$}&\bf{ADM}&\bf{SR-SIM}&\bf{IGM}&\bf{PDSESIM}&\bf{GLD-SR}&\bf{GLD-PFT}\\
\hline
\hline
\bf{Direct}&\bf{SROCC}&0.6867&0.7608&0.8910&0.8432&0.7936&0.9168&0.8903&{0.9223}&0.9090&{0.9206}&0.9095&\bf{0.9242}&\bf{0.9271}&\bf{0.9246}\\
\bf{Average}&\bf{KROCC}&0.5115&0.5831&0.7123&0.6790&0.6221&0.7549&0.7182&{0.7603}&0.7419&{0.7580}&0.7439&\bf{0.7643}&\bf{0.7689}&\bf{0.7639}\\
&\bf{PLCC}&0.7058&0.7619&0.8886&0.8543&0.7961&0.9185&0.8893&{0.9212}&0.9054&{0.9187}&0.9117&\bf{0.9230}&\bf{0.9278}&\bf{0.9273}\\
\hline
\bf{Weighted}&\bf{SROCC}&0.6754&0.8356&0.8908&0.8457&0.7329&0.8971&0.8964&{0.9153}&0.9013&{0.9172}&{0.9155}&\bf{0.9186}&\bf{0.9191}&\bf{0.9208}\\
\bf{Average}&\bf{KROCC}&0.5025&0.6455&0.7104&0.6861&0.5598&0.7326&0.7225&0.7491&0.7343&{0.7541}&{0.7516}&\bf{0.7558}&\bf{0.7575}&\bf{0.7591}\\
&\bf{PLCC}&0.6909&0.8302&0.8828&0.8736&0.7177&0.8972&0.8949&0.9087&0.9001&{0.9119}&{0.9104}&\bf{0.9120}&\bf{0.9193}&\bf{0.9212}\\
\hline
\hline
\end{tabular}
\end{center}
\end{sidewaystable}

\setlength{\tabcolsep}{1.2pt}
\renewcommand{\arraystretch}{0.7}
\begin{sidewaystable}\footnotesize
\begin{center}
\caption{\label{Tab:DBEvalAll}Performance Evaluation in 6 Databases}
\begin{tabular} {ccccccccccccccccc}
\hline
\hline
\bf{Database}& \bf{Measure}&\bf{PSNR}&\bf{SSIM}&\bf{MS-SSIM}&\bf{VIF}&\bf{VSNR}&\bf{MAD}&\bf{IWSSIM}&\bf{FSIM$_c$}&\bf{ADM}&\bf{SR-SIM}&\bf{IGM}&\bf{PDSESIM}&\bf{GLD-SR}&\bf{GLD-PFT}\\
\hline
\hline
\bf{LIVE}&\bf{SROCC}&0.8756&0.9479&0.9513&{0.9636}&0.6481&\bf{0.9669}&0.9566&\bf{0.9645}&0.9460&0.9618&0.9581&\bf{0.9652}&0.9624&0.9631\\
&\bf{KROCC}&0.6865&0.7963&0.8049&0.8282&0.4879&\bf{0.8421}&0.8178&\bf{0.8365}&0.7976&0.8299&0.8250&\bf{0.8378}&{0.8290}&0.8297\\
&\bf{PLCC}&0.8722&0.9449&0.9489&{0.9604}&0.7018&\bf{0.9675}&0.9519&\bf{0.9613}&0.9359&0.9552&0.9567&\bf{0.9616}&0.9506&0.9521\\
&\bf{MAE}&10.483&6.9109&6.6701&6.0952&14.6406&\bf{5.2071}&6.3805&\bf{5.8236}&7.1836&6.3263&{6.0420}&\bf{5.8339}&6.6008&6.4812\\
&\bf{RMSE}&13.3659&8.9437&8.6143&{7.6089}&19.5193&\bf{6.9068}&8.3757&\bf{7.5236}&9.6222&8.0792&7.9557&\bf{7.4929}&8.4907&8.3667\\
\hline									
\bf{CSIQ}&\bf{SROCC}&0.8057&0.8368&0.9132&0.9194&0.8132&\bf{0.9466}&0.9212&0.9309&0.9333&0.9319&{0.9403}&0.9359&\bf{0.9539}&\bf{0.9549}\\
&\bf{KROCC}&0.6078&0.6325&0.7386&0.7532&0.6279&\bf{0.7963}&0.7522&0.7684&0.7710&0.7719&{0.7874}&0.7786&\bf{0.8091}&\bf{0.8108}\\
&\bf{PLCC}&0.7999&0.8144&0.8986&0.9277&0.8016&\bf{0.9505}&0.9142&0.9186&0.9284&0.9241&{0.9267}&0.9244&\bf{0.9506}&\bf{0.9515}\\
&\bf{MAE}&0.1193&0.1144&0.0857&0.0742&0.1138&\bf{0.0631}&0.0789&0.0745&0.0758&0.0734&{0.0698}&0.0703&\bf{0.0619}&\bf{0.0612}\\
&\bf{RMSE}&0.1576&0.1525&0.1168&0.0980&0.1590&\bf{0.0817}&0.1082&0.1060&0.0977&0.1021&{0.1020}&0.1042&\bf{0.0817}&\bf{0.0810}\\
\hline								
\bf{TID2008}&\bf{SROCC}&0.5245&0.8081&0.8528&0.7496&0.7046&0.8340&0.8559&{0.8840}&0.8617&\bf{0.8913}&\bf{0.8901}&\bf{0.8892}&0.8817&0.8849\\
&\bf{KROCC}&0.3696&0.6056&0.6543&0.5863&0.5340&0.6445&0.6636&{0.6991}&0.6842&\bf{0.7149}&\bf{0.7103}&\bf{0.7089}&0.7001&0.7030\\
&\bf{PLCC}&0.5545&0.8061&0.8419&0.8075&0.6507&0.8311&0.8572&{0.8758}&0.8683&\bf{0.8854}&{0.8842}&0.8807&\bf{0.8879}&\bf{0.8909}\\
&\bf{MAE}&0.8918&0.6139&0.5616&0.5837&0.7085&0.5543&0.5245&{0.4868}&0.4883&\bf{0.4543}&{0.4647}&0.4716&\bf{0.4627}&\bf{0.4611}\\
&\bf{RMSE}&1.1168&0.7948&0.7247&0.8007&1.0274&0.7491&0.6915&{0.6482}&0.6664&\bf{0.6246}&{0.6279}&0.6360&\bf{0.6182}&\bf{0.6099}\\
\hline								
\bf{A57}&\bf{SROCC}&0.6189&0.4059&0.8435&0.6224&\bf{0.9355}&0.9023&0.7750&0.9181&0.8725&{0.9295}&0.9006&\bf{0.9299}&\bf{0.9351}&0.9100\\
&\bf{KROCC}&0.4309&0.2780&0.6529&0.4592&\bf{0.8031}&0.7233&0.5880&0.7639&0.6912&{0.7779}&0.7387&\bf{0.7793}&\bf{0.7891}&0.7457\\
&\bf{PLCC}&0.6347&0.4149&0.8394&0.6137&\bf{0.9500}&0.9043&0.7652&{0.9252}&0.8803&0.9247&0.9192&\bf{0.9341}&\bf{0.9354}&0.9217\\
&\bf{MAE}&0.1607&0.1847&0.1119&0.1417&\bf{0.0574}&0.0856&0.1182&0.0794&0.0914&{0.0778}&0.0749&0.0754&\bf{0.0655}&\bf{0.0729}\\
&\bf{RMSE}&0.1899&0.2238&0.1337&0.1957&\bf{0.0770}&0.1051&0.1587&{0.0933}&0.1167&0.0936&0.0968&\bf{0.0877}&\bf{0.0869}&0.0954\\
\hline								
\bf{TOY}&\bf{SROCC}&0.6132&0.7870&0.8874&0.9077&0.8608&\bf{0.9362}&\bf{0.9202}&0.9067&\bf{0.9378}&0.8825&0.8653&0.8963&0.8988&0.9010\\
&\bf{KROCC}&0.4443&0.5922&0.7029&0.7315&0.6745&\bf{0.7823}&\bf{0.7537}&0.7303&\bf{0.7820}&0.6975&0.6734&0.7175&0.7180&0.7210\\
&\bf{PLCC}&0.6498&0.7996&0.8924&0.9136&0.8698&\bf{0.9367}&\bf{0.9246}&0.9072&\bf{0.9415}&0.8870&0.8712&0.8989&0.9060&0.9084\\
&\bf{MAE}&0.7746&0.5649&0.4287&0.4012&0.4599&\bf{0.3493}&\bf{0.3653}&0.4015&\bf{0.3288}&0.4444&0.6149&0.4201&0.4017&0.3961\\
&\bf{RMSE}&0.9517&0.7551&0.5650&0.5096&0.6193&\bf{0.4384}&\bf{0.4775}&0.5275&\bf{0.4225}&0.5782&0.4811&0.5491&0.5302&0.5237\\
\hline								
\bf{IVC}&\bf{SROCC}&0.6825&0.7789&0.8980&0.8964&0.7993&0.9146&0.9125&\bf{0.9293}&0.9026&{0.9265}&0.9027&0.9286&\bf{0.9309}&\bf{0.9336}\\
&\bf{KROCC}&0.5298&0.5939&0.7203&0.7158&0.6053&0.7406&0.7339&{0.7636}&0.7255&{0.7560}&0.7286&\bf{0.7639}&\bf{0.7683}&\bf{0.7731}\\
&\bf{PLCC}&0.7238&0.7915&0.9106&0.9026&0.8026&0.9210&0.9228&{0.9390}&0.8783&{0.9357}&0.9125&\bf{0.9381}&{0.9361}&\bf{0.9391}\\
&\bf{MAE}&0.6288&0.5297&0.3701&0.4050&0.5490&0.3673&0.3684&\bf{0.3277}&0.4471&{0.3404}&0.3702&\bf{0.3228}&{0.3332}&\bf{0.3242}\\
&\bf{RMSE}&0.8582&0.7593&0.5103&0.5276&0.7325&0.4753&0.4704&\bf{0.4203}&0.5885&{0.4306}&0.5036&\bf{0.4246}&{0.4297}&\bf{0.4198}\\
\hline
\hline
\end{tabular}
\end{center}
\end{sidewaystable}







\end{document}